\documentclass[runningheads,orivec]{llncs}

\usepackage[T1]{fontenc}
\usepackage[utf8]{inputenc}

\usepackage{amsmath}

\usepackage{booktabs}
\usepackage{caption}
\usepackage{subcaption}
\usepackage{multirow}
\usepackage{amssymb}
\usepackage{graphicx}
\usepackage{url}
\newcommand{\dn}{$\downarrow$}
\newcommand{\up}{$\uparrow$}

\titlerunning{Attention Head Imbalance in Modality Conflict}

\title{Causal Evidence for Attention Head Imbalance in Modality Conflict Hallucination}

\author{
  Jinrui Jiang\inst{1,2} \and
  Zhangtai Wu\inst{1,2} \and
  Zhen Wu\inst{1,2}\thanks{Corresponding author.} \and
  Xinyu Dai\inst{1,2}
}

\authorrunning{J. Jiang et al.}

\institute{
  National Key Laboratory for Novel Software Technology, Nanjing University, China\\
  \and
  School of Artificial Intelligence, Nanjing University, China\\
  \email{\{jinrui\_jiang, wuzt\}@smail.nju.edu.cn, \{wuz, daixinyu\}@nju.edu.cn}
}

\begin{document}

\maketitle

\begin{abstract}
Modality-conflict hallucination occurs when multimodal large language 
models (MLLMs) prioritize erroneous textual premises over contradictory 
visual evidence. To understand why visual evidence fails to prevail during generation, 
we take a mechanistic perspective and examine which internal components 
drive or resist this failure. We perform head-level causal analysis using path 
patching across five open-source MLLMs and identify two groups of 
attention heads with opposing causal roles: hallucination-driving heads 
and hallucination-resisting heads. We find a consistent asymmetry: driving effects are more broadly distributed 
and carry greater aggregate weight, whereas resisting effects concentrate 
in a small number of high-importance heads.
Ablation experiments further confirm that these groups exert opposing 
effects during generation: distributed driving influence and localized 
resistance together form an imbalanced routing structure that biases 
generation toward the erroneous premise. Motivated by this finding, we propose 
MACI (Modality-conflict-Aware Causal Intervention), a conditional 
intervention that suppresses causally identified hallucination-driving 
heads only when conflict is detected. Across five MLLMs, MACI achieves 
the largest hallucination reduction among compared inference-time 
baselines on the MMMC benchmark with a favorable hallucination-accuracy trade-off, and transfers zero-shot to the SCI-SemanticConflict test.

\end{abstract}

\section{Introduction}
\label{sec:intro}

Modality conflict arises when an erroneous textual premise contradicts visual evidence, such as when a question presupposes the presence of an object that is absent from the image. It constitutes a common source of hallucination in MLLMs: under such conflicts, models often generate answers consistent with the textual premise rather than with the visual evidence, and prior reports indicate hallucination rates exceeding 40\% across the tested models~\cite{zhang2025robust}.

Existing inference-time methods such as 
VCD~\cite{damonlpsg2023vcd}, ICD~\cite{wang2024mitigating}, 
and OPERA~\cite{huang2024opera} mitigate hallucination 
at the output or decoding level, but they provide limited 
insight into the internal causal processes underlying 
modality-conflict failures. Prior work shows that modality-conflict signals are linearly decodable from 
intermediate layers~\cite{nguyen2025challenges}, but this does not explain 
which internal components drive the model toward the erroneous textual premise 
or counteract this tendency.

This motivates a routing-level analysis of how competing textual and visual 
signals are selected and propagated during generation. Because attention 
heads mediate information flow in Transformers~\cite{basu2024understanding,zhang2025crossmodal}, 
they provide a natural unit for testing which components drive or counteract 
modality-conflict hallucination. To move beyond correlational evidence and identify the causal contributions of individual heads, we apply path patching~\cite{vig2020causal,wang2022interpretability}, a mechanistic interpretability technique for causal attribution at the attention-head level.

Applying this head-level causal analysis across five open-source MLLMs, we identify two groups of attention heads with opposing causal roles: hallucination-driving heads, which bias generation toward the erroneous textual premise, and hallucination-resisting heads, which counteract this bias. Crucially, these groups are asymmetric in two respects: driving effects are distributed across more layers and heads and carry greater aggregate weight, whereas resisting effects are concentrated in a small set of high-importance heads. This asymmetry offers causal evidence for one mechanism by which visual 
evidence may fail to prevail under conflict.

As a proof of concept, we propose \textbf{MACI}\footnote{Code will be made 
publicly available upon acceptance.}
(\textbf{M}odality-conflict-\textbf{A}ware \textbf{C}ausal 
\textbf{I}ntervention), a conditional inference-time intervention that leverages 
resisting-head activations to detect conflict and suppresses driving heads 
only when necessary.
Our contributions are as follows:
\begin{itemize}
    \item We provide head-level causal evidence for modality-conflict hallucination, identifying hallucination-driving and hallucination-resisting heads and revealing a consistent twofold asymmetry across five MLLMs.

    \item We validate the identified heads through generation-time ablations 
    (including a random-head control) and show that ablating object-identified 
    driving heads also reduces hallucination on attribute/relation conflicts and the 
    SCI-Semantic\allowbreak{}Conflict test.

    \item We propose MACI, a proof-of-concept conditional intervention that detects conflict from resisting-head activations and suppresses driving heads only when necessary. Across five models, MACI achieves the greatest reduction in hallucinations among the compared inference-time baselines on MMMC and transfers zero-shot to SCI-SemanticConflict.
\end{itemize}

\section{Related Work}
\label{sec:related}

\noindent\textbf{Modality conflict.}
Prior work has identified modality conflict as a distinct failure mode of 
MLLMs~\cite{zhang2025robust,bai2024hallucination}. 
Following MMMC\footnote{\url{https://huggingface.co/datasets/ustc-zhangzm/MMMC}}~\cite{zhang2025robust}, 
which formalizes object, attribute, and relation conflicts, we examine 
erroneous textual premises that contradict ground-truth visual evidence, 
rather than visual-illusion or language-prior 
conflict~\cite{guan2024hallusionbench} or context-parametric 
conflict~\cite{zhu2024unraveling}. To our knowledge, MMMC is the only established
benchmark explicitly designed for this setting; we additionally use the 
SemanticConflict subset of SCI (medium split)\footnote{\url{https://huggingface.co/datasets/sci-benchmark/self-contradictory/viewer/vision-language-4}}~\cite{gao2024dissecting}, 
denoted as SCI-SemanticConflict, whose object-substitution construction 
provides an independent, analogous test. Nguyen et al.~\cite{nguyen2025challenges} 
analyze internal conflict signals, but leave open which components causally 
drive or mitigate modality-conflict hallucination.

\noindent\textbf{Mechanistic interpretability.}
Mechanistic interpretability aims to identify internal components that causally support model behavior. Causal mediation analysis~\cite{vig2020causal}, causal tracing~\cite{meng2022locating}, and path patching
~\cite{wang2022interpretability} have revealed causally important components in language models. In MLLMs, causal tracing has been used to study visual-linguistic information flow~\cite{basu2024understanding,zhang2025crossmodal}, but it has not been applied to attention head-level causal attribution under modality conflict.

\noindent\textbf{Inference-time mitigation.}
VCD, ICD, and OPERA reduce hallucination during decoding
~\cite{damonlpsg2023vcd,wang2024mitigating,huang2024opera}, 
but do not identify internal causal mechanisms. Head-level methods have also been explored~\cite{yang2025modular,wang2025ascd,he2025vhd,qian2025allpaths}: 
Yang et al.~\cite{yang2025modular} identify hallucination heads via modular 
attribution, VHD uses visual-context perturbation, and Intervene-All-Paths 
categorizes image-to-text/text-to-text pathways. Unlike these, MACI targets 
modality conflict, separates driving and resisting 
heads by signed path-patching effects, gating intervention by detected 
conflict.

\section{Causal Analysis of Attention Heads under Modality Conflict}
\label{sec:causal_imbalance}

We use MMMC object conflict as the primary setting because its discrete, unambiguous answer tokens provide a clear basis for log-likelihood-based causal analysis; cross-type generalization is evaluated in Section~\ref{sec:cross_type}.

\begin{figure}[t]
    \centering
    \includegraphics[width=1.0\textwidth]{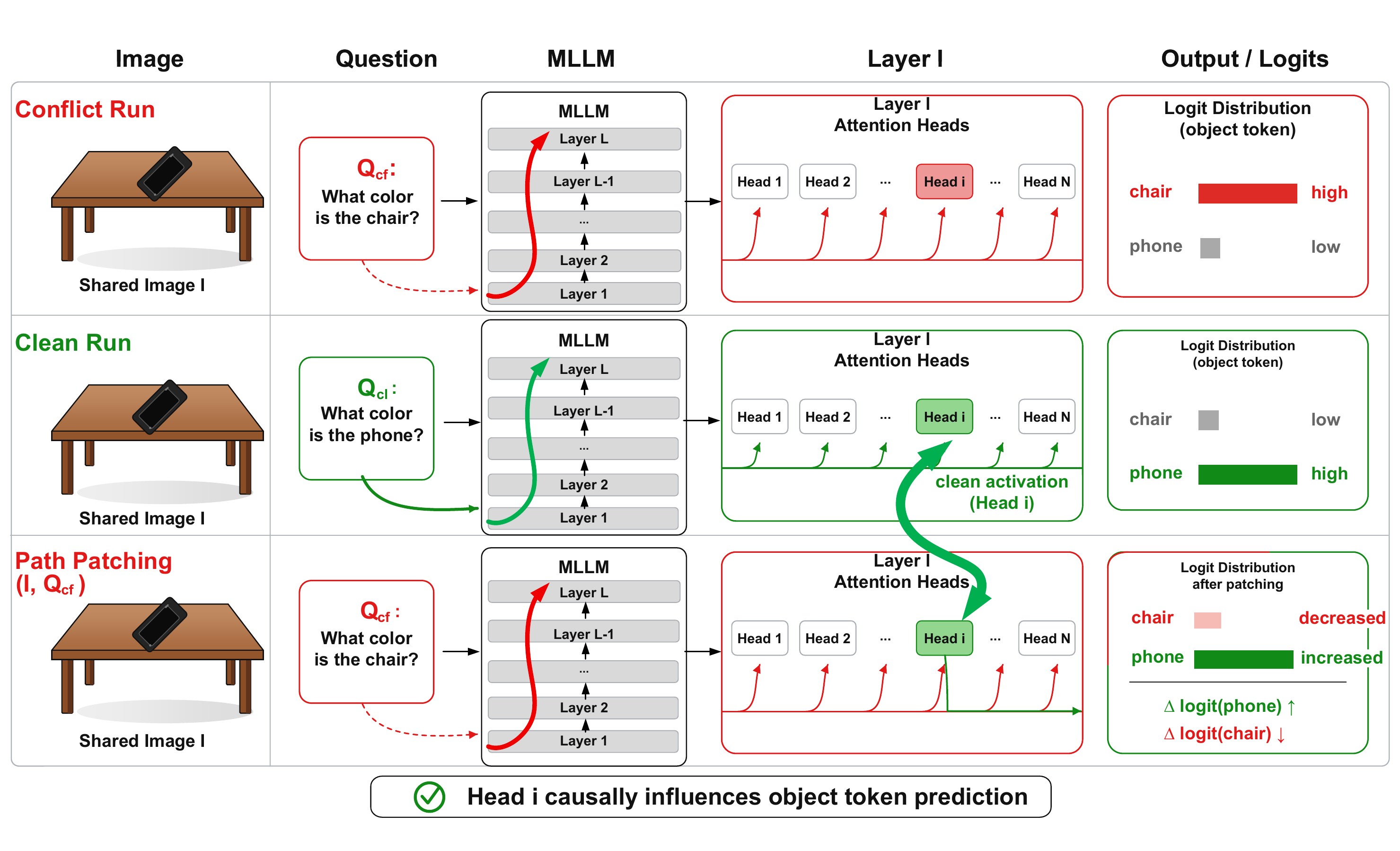}
    \caption{\textbf{Head-level path patching.} 
    \textbf{Top (Conflict run)}: the model is biased toward the erroneous 
    textual premise. \textbf{Middle (Clean run)}: the model identifies the 
    visual evidence given an unbiased query. \textbf{Bottom (Patching)}: 
    substituting head $(l,i)$'s activation with its clean-run counterpart 
    and measuring the change in hallucination advantage indicates whether 
    the head drives or resists hallucination.}
    \label{fig:method}
\end{figure}

\paragraph{Path Patching.}
By temporarily replacing an
attention head's activation with that from a clean run
and observing the resulting change in hallucination
advantage, we can test whether that head causally
influences this tendency. For each sample,
the model runs on a \textbf{conflict} input $(I,Q_{cf})$,
where $Q_{cf}$ carries an erroneous presupposition,
and a \textbf{clean} input $(I,Q_{cl})$, where $Q_{cl}$
is the paired unbiased query from MMMC
(Figure~\ref{fig:method}). Because the two runs share the same image and differ only in the textual 
premise, clean-run activations provide a premise-unbiased reference. 
Patching them into the conflict run tests whether the head amplifies or 
counteracts the hallucination tendency.

\paragraph{Importance Score.}
To quantify the model's tendency to hallucinate, we 
define the \textbf{hallucination advantage} as the 
log-likelihood difference between the hallucinated 
and factual answers:
\begin{equation}
    \mathcal{L}(x) = \log p_\theta(y_h \mid x) - 
    \log p_\theta(y_f \mid x)
    \label{eq:hallucination_advantage}
\end{equation}
where $y_h$ and $y_f$ denote the single-token hallucinated 
and factual object answers, respectively. All head activations used for 
patching are extracted from the prefill forward pass before answer decoding. A positive value 
indicates that the model favors the hallucinated answer implied 
by the erroneous textual premise.

To identify the heads that causally drive or resist hallucination, we patch the clean-run activation of head $(l,i)$ into the conflict run and measure the resulting change in hallucination advantage. Averaged over a prototype set $\mathcal{D}_{proto}$ of 256 training samples, disjoint from the validation and test splits:
\begin{equation}
    \bar{I}_{l,i} = \frac{1}{|\mathcal{D}_{proto}|}
    \sum_{x \in \mathcal{D}_{proto}}
    \left[\mathcal{L}(x_{cf}) -
    \mathcal{L}\!\left(x_{cf}^{(l,i) \leftarrow cl}
    \right)\right]
    \label{eq:importance}
\end{equation}
where $x_{cf}^{(l,i)\leftarrow cl}$ denotes the patched conflict run in which the activation of head $(l,i)$ is replaced by its clean-run counterpart.

\paragraph{Hallucination-driving and -resisting Heads.}
The sign of $\bar{I}_{l,i}$ defines the operational causal role of each head in the conflict run: a positive score indicates that the corresponding head in the original conflict run drives hallucination, whereas a negative score indicates that it resists hallucination. We select the top heads by importance magnitude for each polarity:
\begin{equation}
    \mathcal{H}^+_{k_+} = 
    \underset{(l,i):\,\bar{I}_{l,i}>0}
    {\mathrm{top\text{-}}k_+}\;\bar{I}_{l,i}, \qquad
    \mathcal{H}^-_{k_-} = 
    \underset{(l,i):\,\bar{I}_{l,i}<0}
    {\mathrm{top\text{-}}k_-}\;|\bar{I}_{l,i}|
    \label{eq:heads}
\end{equation}
Heads in $\mathcal{H}^+_{k_+}$ constitute the 
\textbf{hallucination-driving group}, or simply 
\textbf{driving heads}; those in $\mathcal{H}^-_{k_-}$ 
constitute the \textbf{hallucination-resisting group}, or simply 
\textbf{resisting heads}. These designations reflect their operational 
causal roles during modality conflict and do not imply any assumption about the 
underlying signal processed by each head.

\subsection{Identifying Driving and Resisting Heads}
\label{sec:identifying}

Applying the above procedure to five 7B--8B open-source MLLMs 
(Qwen2.5-VL/Qwen3-VL~\cite{bai2025qwen25vl,bai2025qwen3vl}, 
LLaVA/LLaVA-NeXT~\cite{liu2023llava,liu2024llavanext}, and 
InternVL3~\cite{zhu2025internvl3}), spanning dynamic-resolution 
tiling, MLP projection, and cross-attention integration, we find that 
importance scores consistently partition attention heads into two groups 
with opposing signs: positive-scoring heads drive hallucination, whereas 
negative-scoring heads resist it.

\begin{figure}[t]
    \centering
    \includegraphics[width=0.75\textwidth]{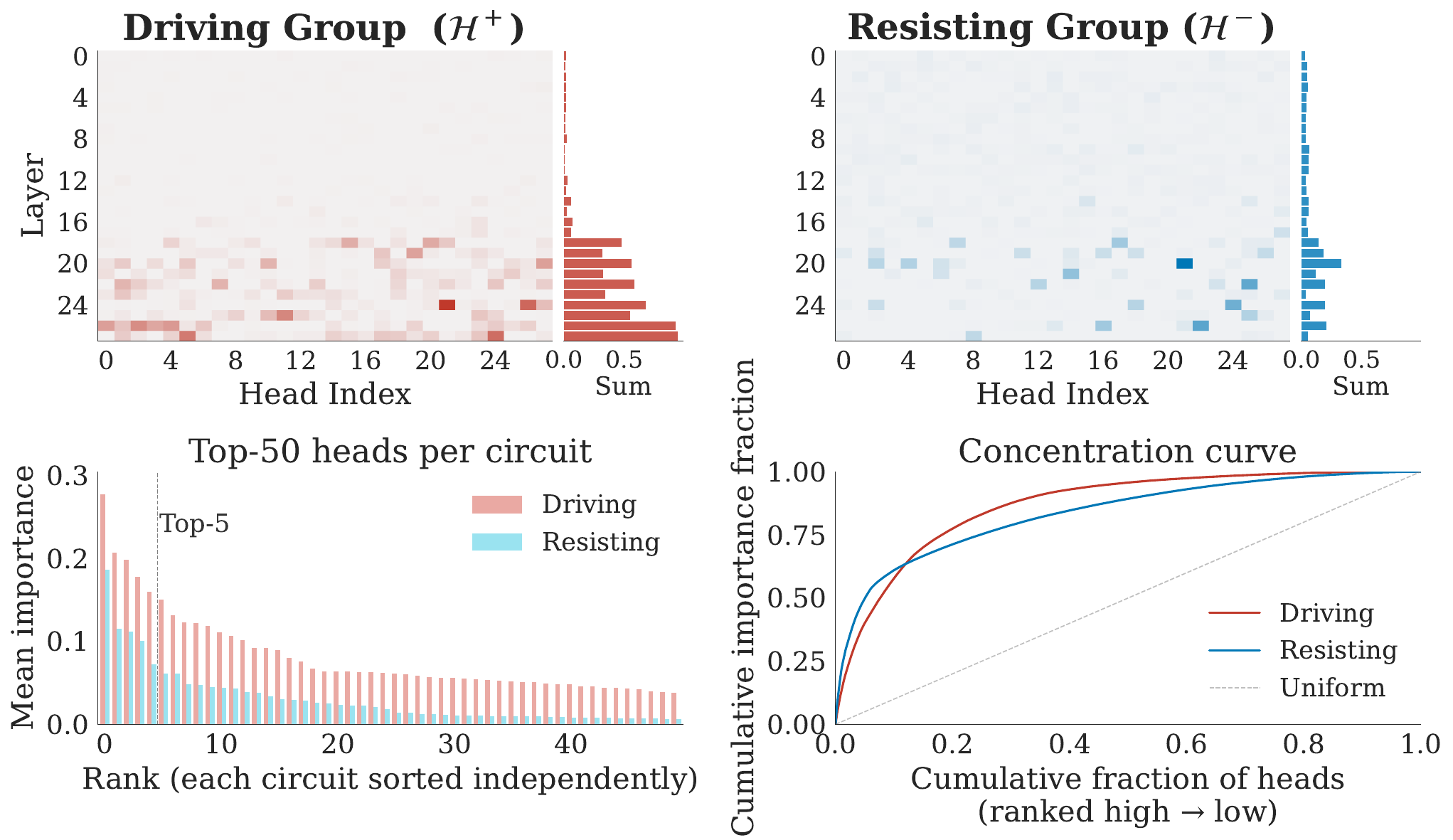}
    \caption{Hallucination-driving ($\mathcal{H}^+$, red) and 
    hallucination-resisting ($\mathcal{H}^-$, blue) heads in 
    Qwen2.5-VL-7B. Top: layer-wise importance and per-layer sums. 
    Bottom: ranked heads and cumulative importance. Results for all 
    models are in Appendix~\ref{app:heatmap}.}
    \label{fig:heatmap}
\end{figure}

Figure~\ref{fig:heatmap} illustrates the driving and resisting heads in Qwen2.5-VL-7B; the same pattern is observed across all five models (Appendix~\ref{app:heatmap}). Both groups are concentrated in middle-to-deep layers, consistent with Nguyen et al.~\cite{nguyen2025challenges}, yet they exhibit complementary head-level asymmetries: the summed positive scores exceed the summed absolute negative scores in all five models (mean per-model ratio: 1.51$\times$), while the top-5 heads account for a larger share of resisting importance than of driving importance (27.2\% vs. 14.0\% on average). This twofold descriptive asymmetry motivates the ablation-based validation in Section~\ref{sec:pruning}, where we examine how the two groups jointly shape generation under natural inference.

\subsection{Causal Validation}
\label{sec:pruning}

To assess whether the importance scores correspond to causal effects under natural inference, in which all heads interact simultaneously, we apply zero ablation during generation and evaluate the hallucination rate rather than the log-likelihood hallucination advantage. We evaluate five conditions on 500 held-out instances: \textbf{Base}, random-head ablation, driving-head ablation, resisting-head ablation, and joint ablation. Ablation suppresses the outputs of selected heads:
\begin{equation}
    \mathbf{a}_{l,i} \leftarrow \mathbf{0},
    \quad \forall (l,i) \in \mathcal{S}
    \label{eq:ablation}
\end{equation}
where $\mathcal{S}$ denotes the heads selected under each ablation condition. For joint ablation, we use equal-size top-$k$ subsets from both polarities to control for group size. Random-head ablation uniformly samples the same number of heads from all $(l,i)$ positions over five seeds, serving as a size-matched control.

\begin{figure}[t]
    \centering
    \includegraphics[width=0.7\textwidth]{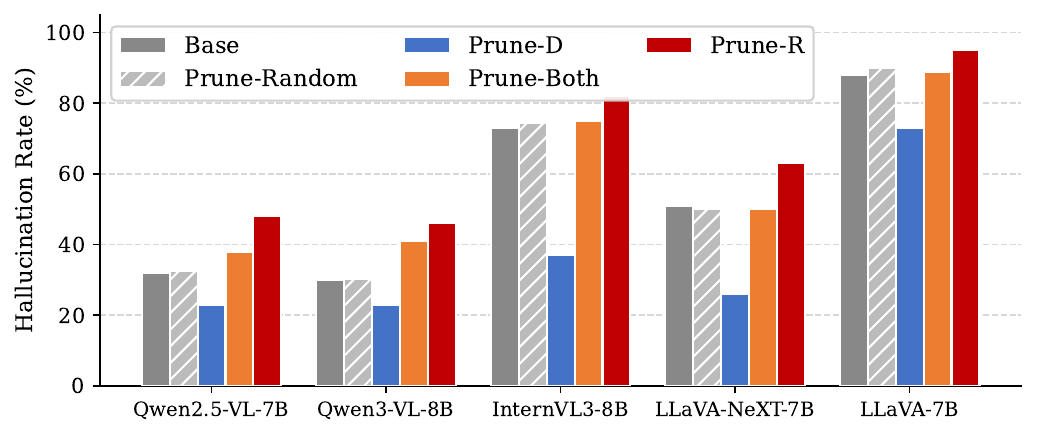}
    \caption{Causal validation by head ablation. Hallucination rate (\%) 
    under Base, Prune-D (driving-head ablation), Prune-R 
    (resisting-head ablation), Prune-Both (joint ablation), and 
    Prune-Random (size-matched random ablation; five-seed average) across 
    five models.}
    \label{fig:pruning}
\end{figure}

Figure~\ref{fig:pruning} shows a consistent pattern across all five models. 
Random-head ablation remains near Base, suggesting that hallucination 
reduction depends on causally selected driving heads rather than generic 
head removal. Driving-head ablation reduces hallucination whereas 
resisting-head ablation increases it; joint ablation yields rates between the 
two single-group ablations, showing that the patching-derived roles remain 
predictive under natural inference. Joint ablation often approaches or exceeds 
Base, suggesting that concentrated resisting heads have substantial behavioral 
effects despite their smaller aggregate importance. Together, these results show that the two groups are not merely 
descriptively asymmetric but exert opposing effects under natural inference.

\subsection{Cross-Type and Cross-Benchmark Generalization}
\label{sec:cross_type}

To assess whether this structure is specific to object conflict or MMMC, 
we examine whether object-identified driving heads retain causal influence 
on attribute, relation conflicts, and SCI-SemanticConflict samples 
(Table~\ref{tab:cross_type}). For SCI-SemanticConflict, which lacks canonical 
answer labels, hallucination is judged by whether the response follows the 
substituted textual premise rather than the image evidence.

Ablating object-identified driving heads reduces hallucination across MMMC 
attribute/relation conflicts and SCI, suggesting that these heads capture 
a broader premise-following tendency beyond object-conflict prompt patterns. The magnitude of transfer varies across models, indicating that architecture, base hallucination rate, or conflict granularity may affect generalization.

\begin{table}[t]
\caption{Cross-type and cross-benchmark transfer of object-identified 
driving heads. Hallucination rate (\%) before and after Prune-D 
(driving-head ablation). SCI denotes the SCI-SemanticConflict subset.}
\label{tab:cross_type}
\centering
\scriptsize
\setlength{\tabcolsep}{1.6pt}
\renewcommand{\arraystretch}{0.86}
\begin{tabular}{llccc}
\toprule
\textbf{Model} & \textbf{Type} & \textbf{Base} & 
\textbf{Prune-D} & \textbf{$\Delta$} \\
\midrule
\multirow{4}{*}{Qwen2.5-VL-7B-Instruct}
  & Obj.  & 28.83 & 22.50 & $-$6.33 \\
  & Attr. & 55.35 & 48.43 & $-$6.92 \\
  & Rel.  & 56.87 & 47.44 & $-$9.43 \\
  & SCI   & 64.00 & 48.60 & $-$15.40 \\
\cmidrule{2-5}
\multirow{4}{*}{Qwen3-VL-8B-Instruct}
  & Obj.  & 30.51 & 25.93 & $-$4.58 \\
  & Attr. & 55.66 & 44.34 & $-$11.32 \\
  & Rel.  & 58.76 & 55.52 & $-$3.24 \\
  & SCI   & 44.60 & 39.10 & $-$5.50 \\
\cmidrule{2-5}
\multirow{4}{*}{LLaVA-NeXT-7B}
  & Obj.  & 44.39 & 22.43 & $-$21.96 \\
  & Attr. & 65.41 & 33.02 & $-$32.39 \\
  & Rel.  & 67.65 & 47.71 & $-$19.94 \\
  & SCI   & 94.80 & 76.10 & $-$18.70 \\
\cmidrule{2-5}
\multirow{4}{*}{InternVL3-8B}
  & Obj.  & 69.03 & 42.87 & $-$26.16 \\
  & Attr. & 84.91 & 59.12 & $-$25.79 \\
  & Rel.  & 80.32 & 62.26 & $-$18.06 \\
  & SCI   & 94.00 & 75.60 & $-$18.40 \\
\cmidrule{2-5}
\multirow{4}{*}{LLaVA-7B}
  & Obj.  & 86.35 & 66.13 & $-$20.22 \\
  & Attr. & 94.34 & 77.36 & $-$16.98 \\
  & Rel.  & 92.45 & 79.51 & $-$12.94 \\
  & SCI   & 99.20 & 95.00 & $-$4.20 \\
\bottomrule
\end{tabular}
\renewcommand{\arraystretch}{1.0}
\end{table}

\section{MACI: A Proof-of-Concept Intervention}
\label{sec:intervention}

\subsection{Conditional Causal Intervention}
\label{sec:method}

Having identified the driving and resisting heads and shown their 
opposing causal effects, we ask whether this mechanistic insight can 
guide targeted intervention. We propose \textbf{MACI} 
(\textbf{M}odality-conflict-\textbf{A}ware \textbf{C}ausal 
\textbf{I}ntervention), a conditional inference-time method that uses 
resisting-head activations to detect conflict and ablates driving heads 
only when necessary. This conditional design avoids the indiscriminate harm of unconditional 
driving-head ablation, applying it only when conflict is detected.

We use resisting-head activations as the conflict-detection signal. Although driving-head activations achieve comparable AUROC (Appendix~\ref{app:judge}), their use would entangle detection with the intervention target. Resisting heads remain untouched, thereby separating detection from action. We average last-token prefill activations of resisting heads into
$\mathbf{h}^{-}(x)=
\frac{1}{|\mathcal{H}^{-}_{k_-}|}
\sum_{(l,i)\in\mathcal{H}^{-}_{k_-}}\mathbf{a}_{l,i}(x)$
and train a Lasso logistic regression probe on the training split. The probe achieves an AUROC of 0.89--0.95 across all five models on object conflict; Appendix~\ref{app:judge} reports zero-shot probe AUROC for attribute and relation conflict.

The intervention is applied conditionally:
\begin{equation}
    \mathbf{a}_{l,i} \leftarrow \mathbf{0}
    \quad\text{if}\quad \hat{y}(x) \geq \tau,
    \forall (l,i) \in \mathcal{H}^+_{k_+}
    \label{eq:conditional}
\end{equation}
where $\hat{y}(x) = \sigma(\mathbf{w}^\top 
\mathbf{h}^-(x) + b)$ and threshold $\tau$ is 
selected on the validation split to maximize F1. 

\subsection{Setup and Results}
\label{sec:results}

\begin{figure}[t]
    \centering
    \includegraphics[width=0.6\textwidth]{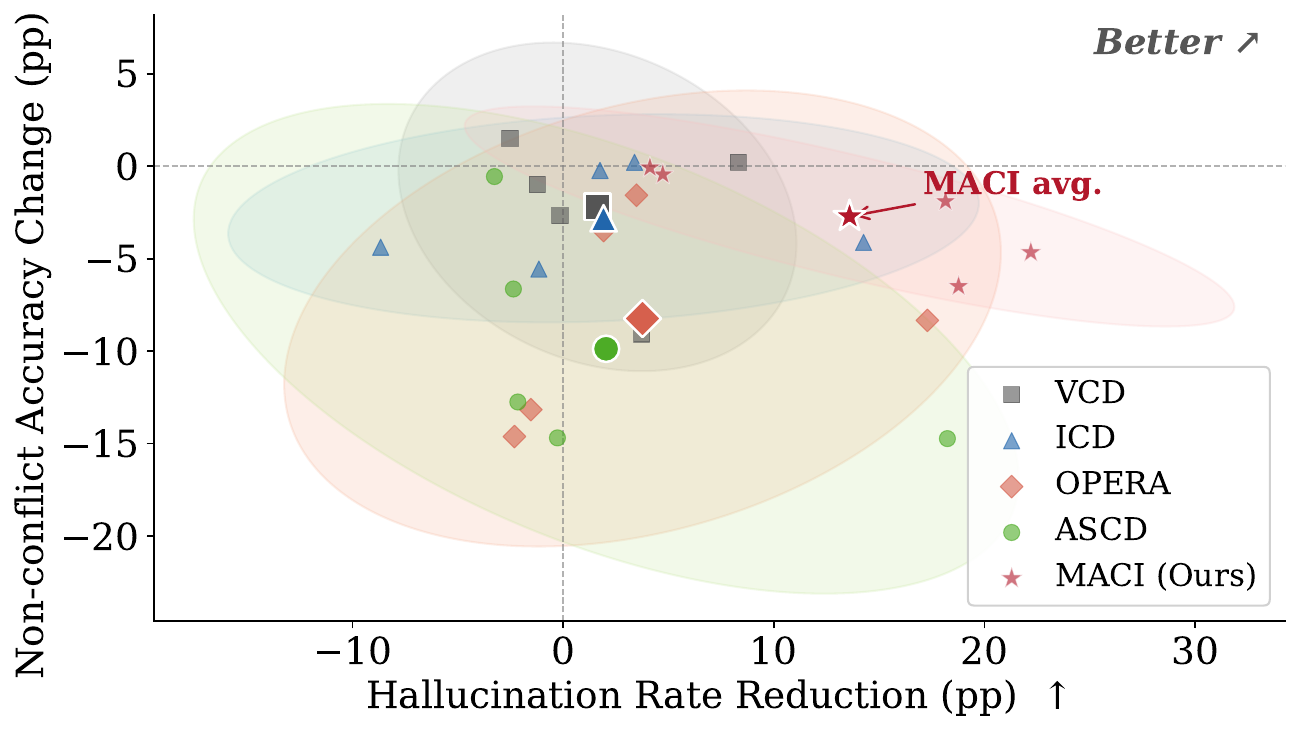}
    \caption{Trade-off between hallucination-rate reduction and 
    non-conflict accuracy change on MMMC. Points are per-model results; 
    ellipses show method-wise spread, and the labeled MACI marker denotes 
    its cross-model average. Upper right is better.}
    \label{fig:tradeoff}
\end{figure}

We compare MACI with \textbf{Base}, VCD~\cite{damonlpsg2023vcd}, 
ICD~\cite{wang2024mitigating}, OPERA~\cite{huang2024opera}, and 
ASCD~\cite{wang2025ascd}. All methods use greedy decoding on the MMMC object conflict 
test split. For MMMC, we report \textbf{Hallucination Rate} 
(answers following the erroneous premise), \textbf{Non-conflict Accuracy} 
(accuracy on paired clean inputs), and \textbf{Overall Rating} 
(mean LLM-as-a-Judge answer-quality score, 0--4~\cite{zheng2023judging}). 
SCI-SemanticConflict is used as a zero-shot cross-benchmark test and reports 
hallucination rate only, since it lacks canonical ground-truth answers. 
Hallucination is judged by DeepSeek-V3.2~\cite{deepseek2025v32} and 
cross-validated with Llama-3.3-70B (Cohen's $\kappa$: 0.784--0.863; 
Appendix~\ref{app:judge}). Results are in Table~\ref{tab:intervention}; 
Figure~\ref{fig:tradeoff} shows the MMMC trade-off.

On MMMC, MACI achieves the largest reduction in hallucination across all five 
models. Non-conflict accuracy remains within 2pp of Base on three models 
(Qwen2.5-VL, Qwen3-VL, LLaVA-NeXT); on InternVL3 and LLaVA, larger accuracy 
drops accompany larger hallucination reductions. These drops may reflect 
residual detector errors or driving heads being less conflict-specific in 
these models. On SCI-SemanticConflict, MACI reduces hallucination on all 
five models (mean $-$7.9pp), supporting zero-shot cross-benchmark transfer 
without re-training the heads, probe, or threshold.

Baselines do not consistently improve upon Base: VCD, ICD, and OPERA show 
model-dependent hallucination effects, while ASCD substantially degrades 
non-conflict accuracy, consistent with attention-statistic head selection 
not separating hallucination-specific from general-purpose heads.

\begin{table}[t]
\caption{Results on MMMC object-conflict and SCI-SemanticConflict. 
For SCI-SemanticConflict, we report only hallucination rate under the same 
substituted-premise criterion.}
\label{tab:intervention}
\centering
\scriptsize
\setlength{\tabcolsep}{1.2pt}
\begin{tabular}{llcccc}
\toprule
\textbf{Model} & \textbf{Method}
  & \multicolumn{3}{c}{\textbf{MMMC}}
  & \textbf{SCI} \\
\cmidrule(lr){3-5}\cmidrule(lr){6-6}
& & \textbf{Hall.\dn}
  & \textbf{Acc.\up}
  & \textbf{Rating\up}
  & \textbf{Hall.\dn} \\
\midrule
\multirow{6}{*}{Qwen2.5-VL-7B-Instruct}
  & Base         & 28.83 & 78.18 & 2.96 & 64.00 \\
  & VCD          & 25.10 & 69.11 & 2.68 & 60.60 \\
  & ICD          & 29.98 & 72.62 & 2.76 & 67.20 \\
  & OPERA        & 30.36 & 65.02 & 2.44 & 66.60 \\
  & ASCD         & 29.10 & 63.49 & 2.49 & 65.80 \\
  & MACI (ours)  & \textbf{24.10} & 77.73 & 2.87 & \textbf{53.80} \\
\cmidrule{2-6}
\multirow{6}{*}{Qwen3-VL-8B-Instruct}
  & Base         & 30.51 & 74.68 & 2.94 & 44.60 \\
  & VCD          & 33.03 & 76.20 & 3.01 & 56.20 \\
  & ICD          & 28.76 & 74.45 & 2.99 & 44.80 \\
  & OPERA        & 32.82 & 60.06 & 2.54 & 43.60 \\
  & ASCD         & 33.77 & 74.12 & 2.94 & 43.40 \\
  & MACI (ours)  & \textbf{26.39} & 74.60 & 2.94 & \textbf{41.20} \\
\cmidrule{2-6}
\multirow{6}{*}{InternVL3-8B}
  & Base         & 69.03 & 73.46 & 2.67 & 94.00 \\
  & VCD          & 60.72 & 73.68 & 2.72 & 92.20 \\
  & ICD          & 54.77 & 69.34 & 2.64 & 91.80 \\
  & OPERA        & 51.74 & 65.12 & 2.45 & 90.40 \\
  & ASCD         & 50.79 & 58.73 & 2.34 & 92.40 \\
  & MACI (ours)  & \textbf{46.83} & 68.80 & 2.47 & \textbf{82.60} \\
\cmidrule{2-6}
\multirow{6}{*}{LLaVA-NeXT-7B}
  & Base         & 44.39 & 71.85 & 2.70 & 94.80 \\
  & VCD          & 45.61 & 70.86 & 2.75 & 94.00 \\
  & ICD          & 53.05 & 67.47 & 2.59 & 95.40 \\
  & OPERA        & 40.91 & 70.28 & 2.86 & 90.80 \\
  & ASCD         & 46.54 & 59.10 & 2.34 & 90.00 \\
  & MACI (ours)  & \textbf{26.24} & 69.95 & 2.63 & \textbf{83.80} \\
\cmidrule{2-6}
\multirow{6}{*}{LLaVA-7B}
  & Base         & 86.35 & 63.39 & 2.52 & 99.20 \\
  & VCD          & 86.50 & 60.72 & 2.38 & 98.40 \\
  & ICD          & 82.96 & 63.60 & 2.41 & 98.80 \\
  & OPERA        & 84.43 & 59.91 & 2.17 & 97.60 \\
  & ASCD         & 88.71 & 56.75 & 2.30 & 99.40 \\
  & MACI (ours)  & \textbf{67.58} & 56.90 & 2.44 & \textbf{95.80} \\
\bottomrule
\end{tabular}
\end{table}

\section{Conclusion}
\label{sec:conclusion}

We investigated the internal causal mechanisms underlying modality conflict in 
MLLMs, identifying hallucination-driving and hallucination-resisting 
attention heads with opposing causal roles. Their twofold asymmetry suggests one mechanism by which visual evidence may 
fail to prevail under conflict: 
distributed driving influence and localized resistance can create an 
imbalanced routing structure that biases generation toward the erroneous textual premise.

As a proof of concept, MACI shows that this structure is actionable. It 
uses resisting-head activations for conflict detection and suppresses 
driving heads only when necessary, reducing hallucination with a favorable hallucination-accuracy trade-off on MMMC and transferring zero-shot to 
SCI-SemanticConflict.

Several limitations remain: component identification relies on object-conflict data and the prefill stage, and the probe requires labeled samples, although resisting-head activations support zero-shot conflict detection for attribute and relation conflicts (Appendix~\ref{app:judge}). The evaluation focuses on MMMC, the dedicated benchmark for this setting, with SCI-SemanticConflict serving as an independent cross-benchmark test; extending the approach to settings in which the visual input itself may be misleading, rather than serving as ground-truth evidence, remains future work. Future work may also further leverage resisting-head signals, better balance the two influences during training, and use resisting-head activations as lightweight visual-fidelity signals.


\bibliography{custom}
\appendix

\section{Judge Reliability and Probe Detection}
\label{app:judge}

Hall. labels premise following, not correctness: erroneous-premise following 
on MMMC and substituted-premise following on SCI. Llama-3.3-70B gives 
$\kappa=0.784$--0.863 on MMMC; manual audits show 89\%/91\% agreement on 
100 MMMC/
SCI outputs. 

\noindent
\begin{minipage}[t]{0.48\columnwidth}
\captionof{table}{MMMC hallucination rate (\%) under Llama-3.3-70B.
Cohen's $\kappa$: 0.784--0.863.}
\label{tab:cross_validation}
\centering
\scriptsize
\setlength{\tabcolsep}{2pt}
\begin{tabular}{llcc}
\toprule
\textbf{Model} & \textbf{Method} & \textbf{Hall.\dn} &
\textbf{Acc.\up} \\
\midrule
\multirow{2}{*}{Qwen2.5-VL} & Base & 25.73 & 75.04 \\
  & MACI & 22.88 & 75.29 \\
\multirow{2}{*}{Qwen3-VL}   & Base & 27.46 & 72.39 \\
  & MACI & 24.03 & 70.18 \\
\multirow{2}{*}{InternVL3}  & Base & 66.82 & 71.62 \\
  & MACI & 42.33 & 68.34 \\
\multirow{2}{*}{LLaVA-NeXT} & Base & 41.65 & 68.12 \\
  & MACI & 23.04 & 65.14 \\
\multirow{2}{*}{LLaVA}      & Base & 85.13 & 62.09 \\
  & MACI & 63.46 & 60.64 \\
\bottomrule
\end{tabular}
\end{minipage}
\hfill
\begin{minipage}[t]{0.48\columnwidth}
\captionof{table}{MMMC conflict detection AUROC for resisting-head 
and driving-head probes.}
\label{tab:probe_full}
\centering
\scriptsize
\setlength{\tabcolsep}{2pt}
\begin{tabular}{llccc}
\toprule
\textbf{Model} & \textbf{Heads} &
\textbf{Obj.} & \textbf{Attr.} & \textbf{Rel.} \\
\midrule
\multirow{2}{*}{Qwen2.5-VL}
  & Res. & 0.9484 & 0.8871 & 0.7622 \\
  & Drv. & 0.9253 & 0.8490 & 0.7236 \\
\cmidrule{2-5}
\multirow{2}{*}{Qwen3-VL}
  & Res. & 0.9176 & 0.8638 & 0.7430 \\
  & Drv. & 0.9187 & 0.8651 & 0.7572 \\
\cmidrule{2-5}
\multirow{2}{*}{InternVL3}
  & Res. & 0.8970 & 0.8151 & 0.7316 \\
  & Drv. & 0.9080 & 0.8378 & 0.7631 \\
\cmidrule{2-5}
\multirow{2}{*}{LLaVA-NeXT}
  & Res. & 0.9296 & 0.8128 & 0.7100 \\
  & Drv. & 0.9269 & 0.7474 & 0.6971 \\
\cmidrule{2-5}
\multirow{2}{*}{LLaVA}
  & Res. & 0.9137 & 0.7717 & 0.6953 \\
  & Drv. & 0.8826 & 0.7364 & 0.6643 \\
\bottomrule
\end{tabular}
\end{minipage}

\section{Head Importance Distributions}
\label{app:heatmap}

\begin{figure}[h]
    \centering
    \captionsetup[subfigure]{skip=1pt}
    \begin{subfigure}[t]{0.485\textwidth}
        \centering
        \includegraphics[width=\textwidth]{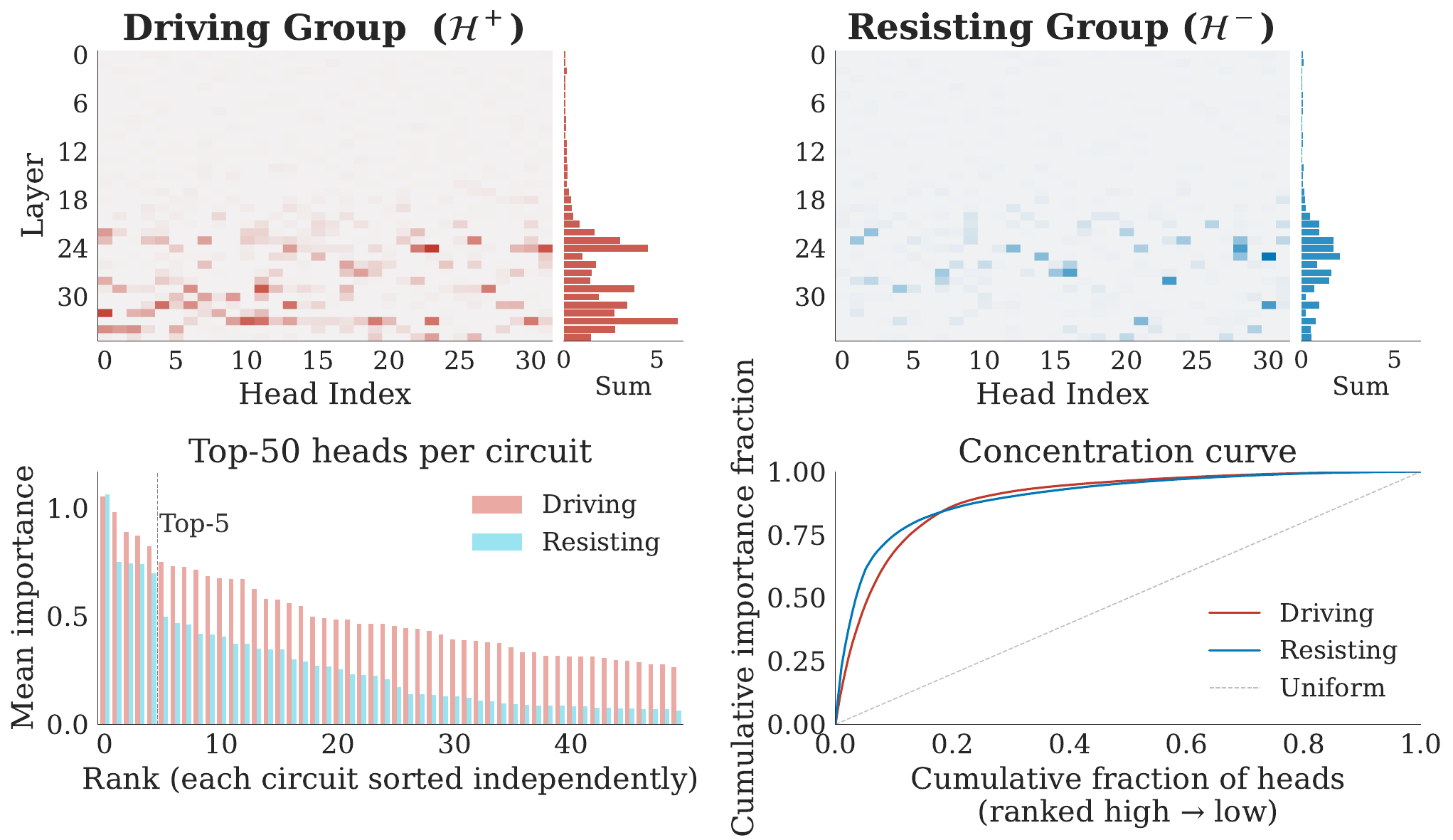}
        \caption{Qwen3-VL-8B}
    \end{subfigure}\hfill%
    \begin{subfigure}[t]{0.485\textwidth}
        \centering
        \includegraphics[width=\textwidth]{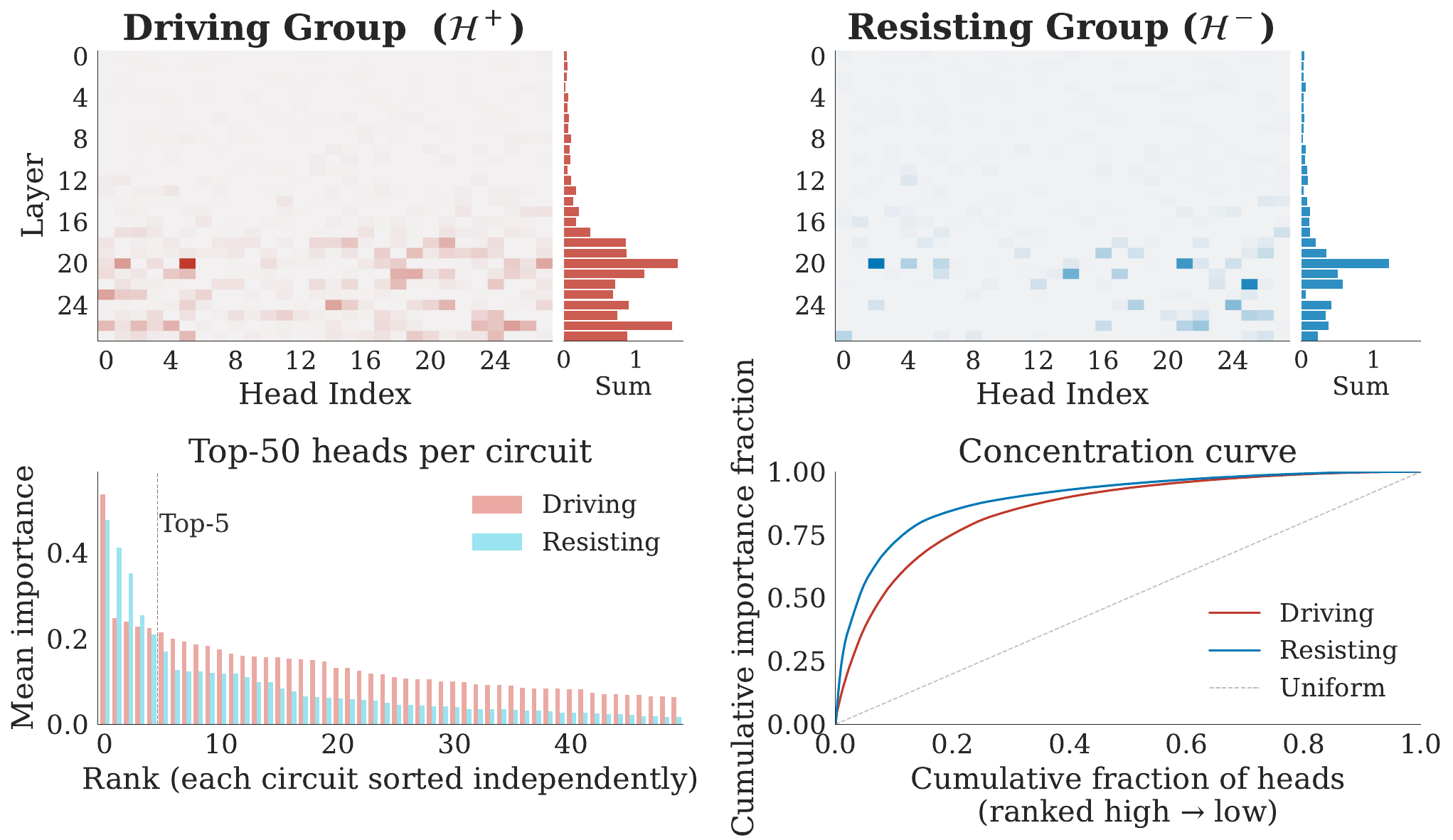}
        \caption{InternVL3-8B}
    \end{subfigure}

    \vspace{-0.4em}

    \begin{subfigure}[t]{0.5\textwidth}
        \centering
        \includegraphics[width=\textwidth]{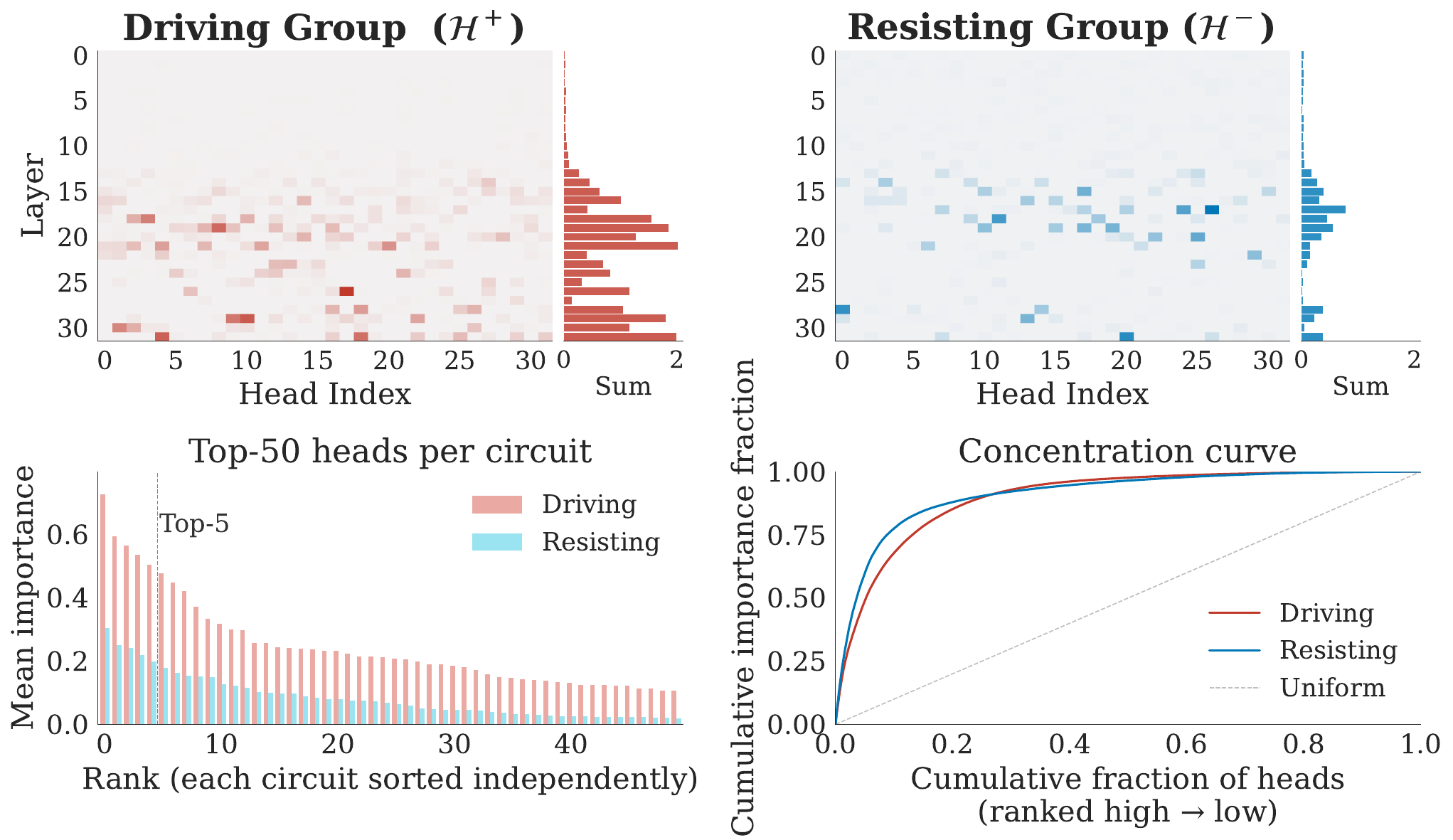}
        \caption{LLaVA-NeXT-7B}
    \end{subfigure}\hfill%
    \begin{subfigure}[t]{0.5\textwidth}
        \centering
        \includegraphics[width=\textwidth]{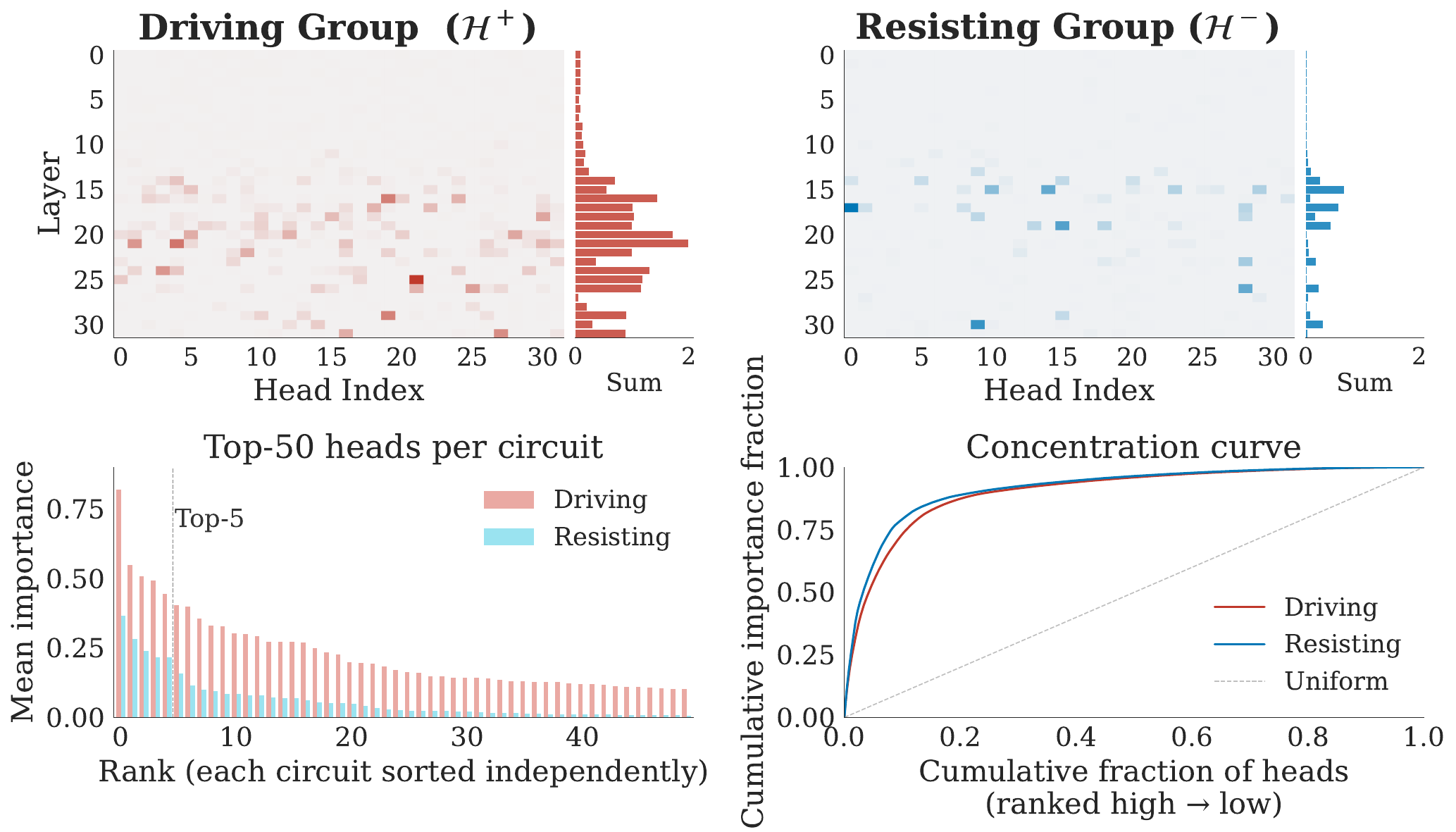}
        \caption{LLaVA-7B}
    \end{subfigure}
    \caption{Head importance distributions for the remaining four models.}
    \label{fig:heatmap_all}
\end{figure}

\section{Implementation Details}
\label{app:implementation}

For MACI, $k_+=(30,30,47,50,64)$ for Qwen2.5-VL, Qwen3-VL, 
InternVL3, LLaVA-NeXT, and LLaVA, respectively, and $k_-=40$ for all 
models. Joint ablation uses equal-size top-$k$ subsets with $k=30$; 
random-head ablation matches the driving-head ablation size and is averaged 
over five seeds. Cross-type and SCI transfer reuse the object-identified 
$\mathcal{H}^+_{k_+}$ and zero-ablation protocol without re-identification. We use zero ablation because mean activation replacement produced unstable 
generations in autoregressive decoding.
Split-half prototype sets yield 79.6\%/64.7\% top-30 overlap for 
driving/resisting heads on average, indicating stable head identification. No test 
instances are used for head identification, probe training, or threshold 
selection. Baselines use official or recommended hyperparameters.

\paragraph{Sensitivity.}
Sweeping $k_+$ over 0.5\%--15\% of all attention heads on the validation 
split shows the expected suppression--preservation trade-off: smaller values 
weaken hallucination suppression, while larger values increase non-conflict 
accuracy loss. For $k_-$, validation AUROC is lower with too few resisting 
heads and also drops when too many low-importance heads are included; we 
therefore use a single intermediate value $k_-=40$ for all models. These hyperparameters are validation-selected rather than test-tuned.
\end{document}